\documentclass[runningheads]{llncs}
\usepackage[T1]{fontenc}
\usepackage{graphicx,verbatim}
\usepackage{multirow, multicol}
\usepackage{booktabs}
\usepackage[table]{xcolor}
\usepackage{amssymb}
\usepackage{amsmath}
\usepackage{marvosym}
\usepackage[hidelinks]{hyperref}
\usepackage{color}

\begin{document}
\title{Adaptively Distilled ControlNet: Accelerated Training and Superior Sampling for Medical Image Synthesis}
\titlerunning{Adaptively Distilled ControlNet}
\author{
    Kunpeng Qiu\inst{1,2} \and
    Zhiying Zhou\inst{1,2} \and
    Yongxin Guo\inst{1,2,3}\textsuperscript{(\Letter)}
}
%index{Qiu, Kunpeng}
%index{Zhou, Zhiying}
%index{Guo, Yongxin}
\authorrunning{Qiu et al.}
\institute{
    National University of Singapore, Singapore, Singapore \and
    National University of Singapore Suzhou Research Institute, Suzhou, China \and
    City University of Hong Kong, Hong Kong, China \\
    \email{kunpeng\_qiu@u.nus.edu}, \email{ elezzy@nus.edu.sg}, \email{ yongxin.guo@cityu.edu.hk}
}
\maketitle              % typeset the header of the contribution
\begin{abstract}
Medical image annotation is constrained by privacy concerns and labor-intensive labeling, significantly limiting the performance and generalization of segmentation models. While mask-controllable diffusion models excel in synthesis, they struggle with precise lesion-mask alignment. We propose \textbf{Adaptively Distilled ControlNet}, a task-agnostic framework that accelerates training and optimization through dual-model distillation. Specifically, during training, a teacher model, conditioned on mask-image pairs, regularizes a mask-only student model via predicted noise alignment in parameter space, further enhanced by adaptive regularization based on lesion-background ratios. During sampling, only the student model is used, enabling privacy-preserving medical image generation. Comprehensive evaluations on two distinct medical datasets demonstrate state-of-the-art performance: TransUNet improves mDice/mIoU by 2.4\%/4.2\% on KiTS19, while SANet achieves 2.6\%/3.5\% gains on Polyps, highlighting its effectiveness and superiority. Code is available at \url{https://github.com/Qiukunpeng/ADC}.

\keywords{Diffusion models \and Medical Image Synthesis \and Medical Image Segmentation.}
\end{abstract}
\section{Introduction}
In medical image analysis, large, accurately annotated datasets are essential for high-performance segmentation \cite{DiffuMask}. Despite the rapid progress in deep learning \cite{SANet,Polyp-PVT,nnUNet,Transunet,Rethinking}, the high cost of acquiring annotated medical images, coupled with privacy and copyright constraints \cite{DiffuseExpand,ArSDM,Learn,Enhancing,SV-DRR}, hinders the full potential of segmentation models.

To mitigate data scarcity issue, diffusion models \cite{DDPM,SD} have emerged as a leading paradigm for synthetic data generation, offering both training stability and high-fidelity image synthesis. Several existing approaches leverage lesion-free images \cite{PolypGen} to synthesize abnormal samples; however, these methods fail to fully address privacy concerns. In contrast, mask-controllable synthesis eliminates the need for costly manual annotations and ethical constraints while providing a more accessible and streamlined framework, making it a compelling alternative for broader adoption \cite{ArSDM,DiffuseExpand,Siamese-Diffusion}. Regardless of the approach, precise lesion-mask alignment remains a notorious challenge in existing methods \cite{T2I,ControlNet,ControlNetPlus,ArSDM}. In this work, we advance the mask-controllable synthesis paradigm to generate high-quality synthetic medical images, specifically tackling lesion alignment limitations to enhance downstream segmentation performance.

To address this, studies \cite{ControlNetPlus,ArSDM} have embedded pretrained segmentation models within diffusion frameworks to provide iterative feedback, refining noise prediction. However, their reliance on pretrained segmentation models renders these methods task-specific and may introduce inherent biases into synthetic data. In a related effort, \cite{ArSDM} introduces adaptive weighting to enhance lesion representation, yet the disproportionately low weight assigned to lesion-free regions impairs learning, leading to degraded image fidelity even after extensive training.

To overcome these limitations, we propose the \textbf{Adaptively Distilled ControlNet}, a novel field distillation framework \cite{DistillationClassifier,Consistency}. Our approach leverages the regularization property of controllable diffusion models \cite{Classifier,Classifier-free}, where conditional inputs act as implicit regularizers to ensure stable optimization and enhanced image quality \cite{Siamese-Diffusion}. Specifically, we adopt a teacher-student paradigm, where the teacher model—conditioned on mask-image pairs—regularizes the noise prediction of the student model, which is conditioned only on masks. A shared forward noise addition process enables a dual-diffusion decoder architecture. Furthermore, an adaptive weight distillation strategy reinforces lesion representation while preserving distributional fidelity. During sampling, the student model runs at ControlNet \cite{ControlNet} speed while ensuring diversity and scalability without extra image conditions. 

Our contributions are summarized as follows: (1) We introduce Adaptively Distilled ControlNet, which significantly accelerates training convergence and data fitting. Moreover, its task-agnostic nature allows seamless adaptation to diverse datasets and modalities without requiring modifications to the model architecture. (2) We propose Adaptive Distillation Loss, which substantially enhances lesion-mask alignment in synthetic images, generating high-quality training data for segmentation models. This ensures superior performance and generalization in downstream segmentation tasks. (3) Extensive experiments demonstrate that our method surpasses existing approaches in both image fidelity and segmentation accuracy. Specifically, TransUNet achieves 2.4\% mDice and 4.2\% mIoU improvements on the KiTS19 dataset, while SANet attains 2.6\% mDice and 3.5\% mIoU gains on Polyps, underscoring the efficacy of our approach.

\begin{figure}[!t]
\includegraphics[width=\textwidth]{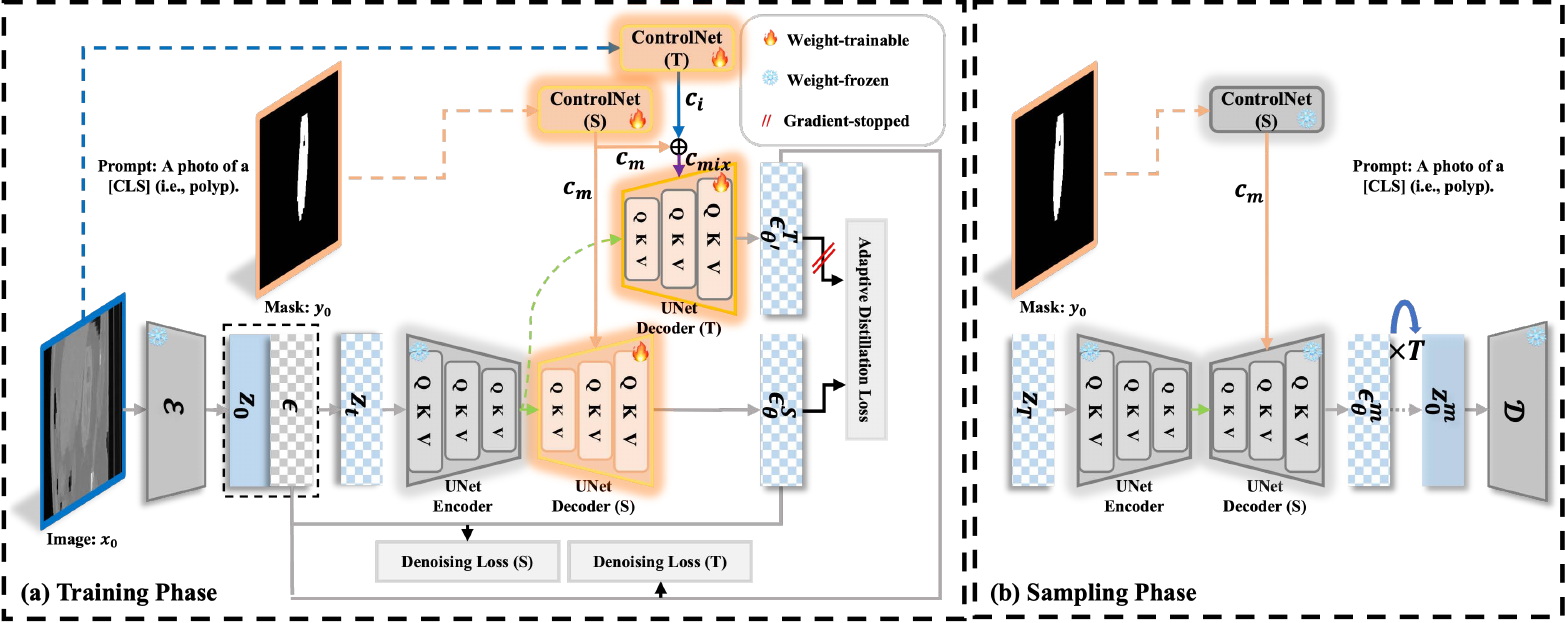}
\caption{(a) Illustration of our method during the training phase. (b) During sampling, only the student model is utilized with arbitrary masks.} \label{fig1}
\end{figure}

\section{Preliminary}
Diffusion models \cite{DDPM,Score-based} formalize data generation through two coupled chains: a destructive forward process that gradually corrupts data with Gaussian noise, and a learned reverse process that iteratively recovers the original signal. Following the standard variance-preserving formulation \cite{DDPM}, the denoising network $\epsilon_\theta(x_t,t)$ directly predicts the noise, reducing the training objective to:
\begin{equation}\label{eq1}
\mathcal{L}_{\text{simple}} = \mathbb{E}_{x_t,t,\epsilon}\left[ \|\epsilon_\theta(x_t,t) - \epsilon\|_2^2 \right],
\end{equation}
where $t \sim \mathcal{U}\{1,T\}$ and $x_t$ is the noisy image.

Stable Diffusion \cite{SD} refines this framework through latent space optimization. A pretrained VAE \cite{VQVAE} encoder $\mathcal{E}$ maps images $x_0$ into compact latent representations $z_0 = \mathcal{E}(x_0)$, facilitating diffusion in a reduced-dimensional space. Various extensions \cite{T2I,ControlNet,ControlNetPlus} of this model enable conditional generation via text prompts $c_t$ and task-specific control signals $c_f$, allowing for more precise content modulation. The generalized training objective is expressed as:
\begin{equation}\label{eq2}
\mathcal{L}_{\text{cond}} = \mathbb{E}_{z_t,t,c_t,c_f,\epsilon}\left[ \|\epsilon_\theta(z_t,t,c_t,c_f) - \epsilon\|_2^2 \right].
\end{equation}

\section{Methodology}
\subsection{Architecture of Adaptively Distilled ControlNet}
Building upon the established ControlNet framework \cite{ControlNet}, we propose a distilled dual-branch diffusion architecture with shared latent projection, as illustrated in Fig.~\ref{fig1}(a). The frozen VAE \cite{VQVAE} encoder $\mathcal{E}$ establishes a deterministic mapping $\mathcal{E}:x_0 \mapsto z_0$ through latent space embedding, where $x_0$ denotes the input image and $z_0$ its latent representation. The student branch (S) ingests conditional masks through a dedicated ControlNet (S) module, generating encoded mask features $c_m$ that integrate with the student diffusion U-Net Decoder (S) through feature injection for noise prediction $\epsilon_{\theta}^{S}$.

The teacher branch (T) processes the paired image through a parallel ControlNet (T) to extract encoded image features $c_i$. These image features are fused with the corresponding mask features $c_m$ through element-wise summation:
\begin{equation}\label{eq3}
c_{\text{mix}} = c_i + c_m.
\end{equation}
This fused representation $c_{\text{mix}}$ propagates through the teacher’s diffusion U-Net decoder (T) to predict the noise $\epsilon_{\theta^{\prime}}^{T}$. By sharing the forward process between the student and teacher branches, the architecture employs a unified latent space projection and diffusion U-Net encoder, significantly optimizing memory efficiency. The composite objective function integrates the following components:
\begin{equation}\label{eq4}
\mathcal{L} = \underbrace{\mathcal{L}_S + \mathcal{L}_T}_{\text{Denoising Objectives}} + \underbrace{\mathcal{L}_{\text{Ada}}}_{\text{Distillation Regularizer}},
\end{equation}
with Denoising Objectives defined as:
\begin{equation}\label{eq5}
\begin{aligned}
\mathcal{L}_S &= \mathbb{E}_{z_t,t,c_t,c_m,\epsilon} \left[ \|\epsilon_{\theta}(z_t, t, c_t, c_m) - \epsilon\|_2^2 \right], \\
\mathcal{L}_T &= \mathbb{E}_{z_t,c_t,c_{mix},t,\epsilon} \left[ \|\epsilon_{\theta^{\prime}}(z_t, t, c_t, c_{\text{mix}}) - \epsilon\|_2^2 \right],
\end{aligned}
\end{equation}
where $\theta$ and $\theta^{\prime}$, as in ControlNet \cite{ControlNet}, are both initialized with the parameters of a pretrained diffusion model, denote mutually independent parameters for each branch, and are optimized separately during training. Meanwhile, $\epsilon \sim \mathcal{N}(0,I)$ ensures stochastic consistency.

\begin{figure}[!t]
\includegraphics[width=\textwidth]{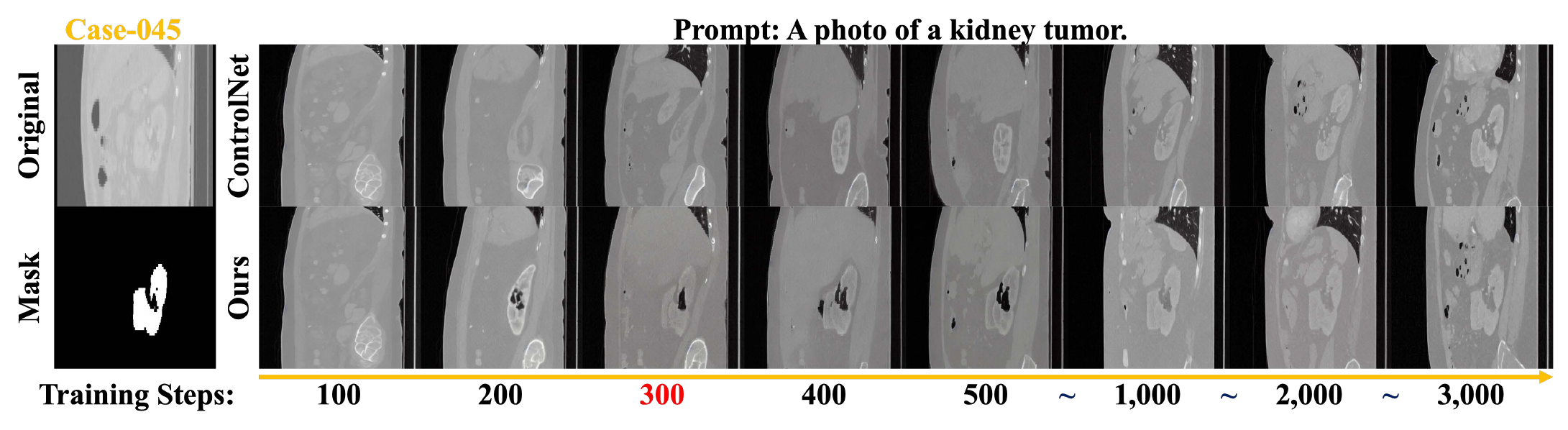}
\caption{Visualizing the difference between ControlNet and our method in training convergence and data fitting.} \label{fig2}
\end{figure}

\begin{table*}[t]
\centering
\caption{Comparison of synthetic medical image quality generated by each method.}
\renewcommand\tabcolsep{0.9pt}
{\fontsize{8}{10}\selectfont
\begin{tabular}{@{}lccccc|ccc@{}}
\toprule
\multirow{2}{*}{Metrics}
& \multicolumn{5}{c|}{Polyps} & \multicolumn{3}{c}{KiTS19} 
\\ \cmidrule(l){2-9} 
&SinGAN & ArSDM & T2I-Adapter & ControlNet & Ours  & T2I-Adapter & ControlNet & Ours\\
\midrule
FID ($\downarrow$) & 103.142 & 98.085 & 150.546 & \textbf{65.609} & 66.587 & 92.717 & \textbf{69.240} & 70.786 \\ 
CLIP-I ($\uparrow$) & 0.851 & 0.845 & 0.874 & 0.884 & \textbf{0.901} & 0.814 & 0.833 & \textbf{0.839} \\
\bottomrule
\end{tabular}}
\label{table1}
\end{table*}

During sampling, as shown in Fig.~\ref{fig1}(b), medical images are generated using the student branch with arbitrary masks at the same speed as ControlNet \cite{ControlNet}.

\subsection{Adaptive Distillation Loss}
The spatial alignment between synthesized lesion regions and their corresponding masks is critical for downstream segmentation tasks. However, the severe lesion-background imbalance in medical image synthesis often leads to the underrepresentation of lesion regions. To address this issue, we propose a spatially adaptive distillation mechanism that enables the teacher model to dynamically modulate the regularization intensity for the student model, thereby emphasizing the learning of lesion-specific morphological features in the student model.

\begin{figure}[!t]
\includegraphics[width=\textwidth]{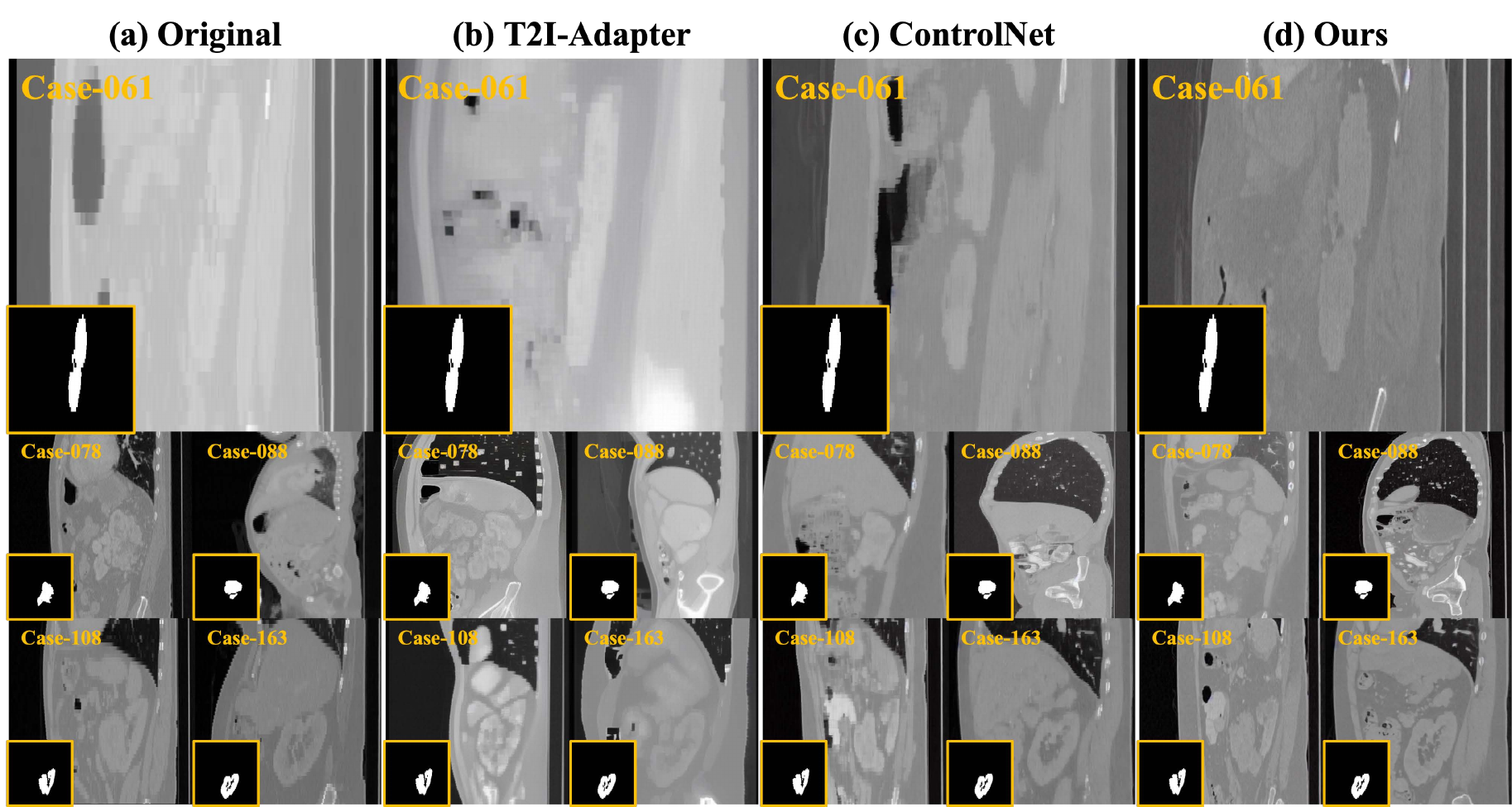}
\caption{Examples of real and synthetic kidney tumor images generated by each method.} \label{fig3}
\end{figure}

\begin{table*}[t]
\centering
\caption{Comparisons of different methods applied on  tumor segmentation baselines.}
\renewcommand\tabcolsep{0.9pt}
{\fontsize{8}{10}\selectfont
\begin{tabular}{@{}lcccc|cccc@{}}
\toprule
\multirow{2}{*}{Methods}
& \multicolumn{4}{c|}{TransUNet} & \multicolumn{4}{c}{nnUNet} 
\\ \cmidrule(l){2-9} 
&mDice &mIoU &Accuracy &Recall &mDice &mIoU &Accuracy &Recall\\
\midrule
Real Dataset & 92.8 & 86.9 & 98.6 & 91.5 & 96.5 & 93.4 & 99.3 & 96.4 \\ 
+Copy-Paste & 93.3 & 87.7 & 98.7 & 91.5 & 96.5 & 93.6 & 99.3 & 96.0 \\
+T2I-Adapter & 94.5 & 89.9 & \textbf{99.0} & 92.6 & 96.3 & 93.6 & \textbf{99.8} & 95.8 \\
+ControlNet & 94.6 & 90.0 & \textbf{99.0} & \textbf{93.9} & 96.1 & 93.2 & \textbf{99.8} & 95.8\\
+Ours & \textbf{95.2} & \textbf{91.1} & \textbf{99.0} & 93.8 & \textbf{97.9} & \textbf{96.0} & 99.6 & \textbf{97.8} \\
\bottomrule
\end{tabular}}
\label{table2}
\end{table*}

Unlike previous approaches that apply reweighting techniques to denoising losses \cite{ArSDM}, our method introduces lesion-aware attention through dual-stream gradient modulation, effectively addressing the lesion-background imbalance. The adaptive weight $w_{\text{Ada}}$ is derived from the mask statistics, with distinct weights assigned to lesion and lesion-free regions:
\begin{equation}\label{eq7}
w_{\text{Ada}} = \begin{cases}
\frac{N_{\text{lesion-free}}}{N_{\text{total}}}, & \text{for lesion regions} \\
\frac{N_{\text{lesion}}}{N_{\text{total}}}, & \text{otherwise}
\end{cases}
\end{equation}
where $N_{\text{lesion}}$ and $N_{\text{lesion-free}}$ denote pixel counts for respective regions, and $N_{\text{total}} = H \times W$ represents the total number of pixels in the image. These weights are normalized to form a spatially adaptive $W \times H$ weight matrix. The final adaptive distillation loss is formulated as:
\begin{equation}\label{eq8}
\mathcal{L}_{\text{Ada}} = \mathbb{E}_{z_t,t} \left[w_{\text{Ada}} \cdot \|\epsilon_{\theta}^S  - \text{sg}(\epsilon_{\theta^{\prime}}^T)\|_2^2 \right],
\end{equation}
where $\text{sg}(\cdot)$ indicates stop-gradient operation.

\begin{figure}[!t]
\includegraphics[width=\textwidth]{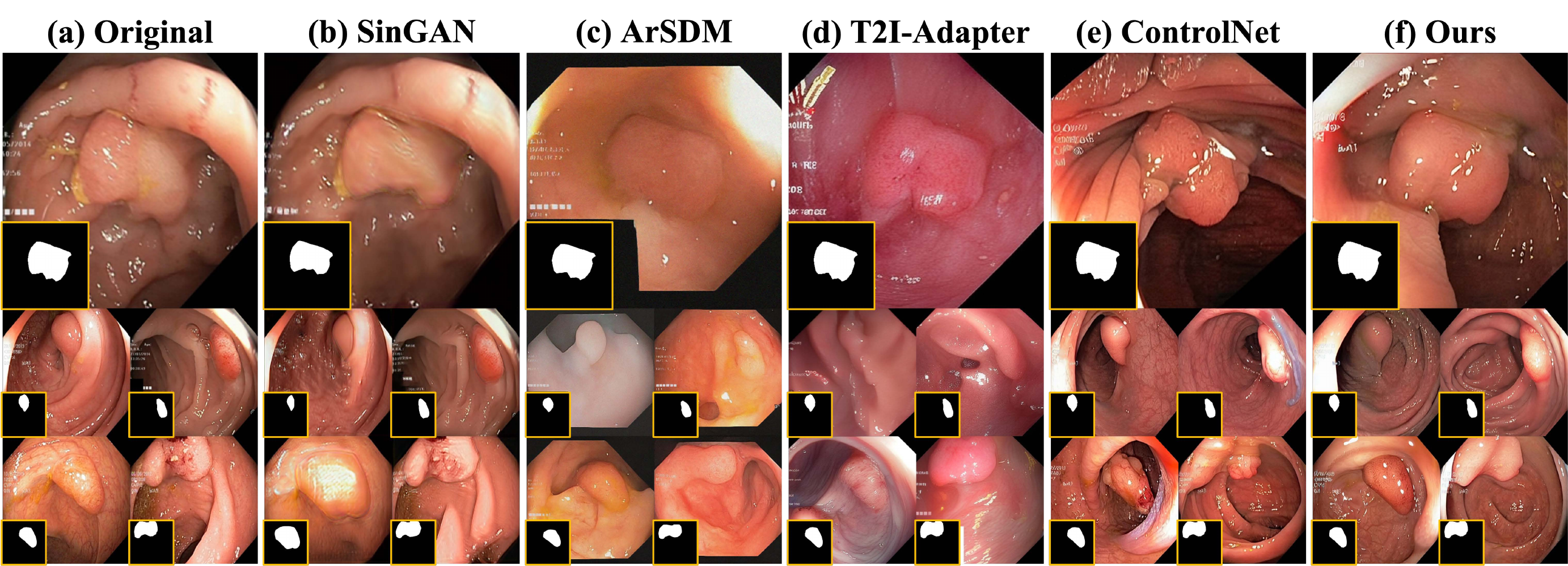}
\caption{Examples of real and synthetic polyp images generated by each method.} \label{fig4}
\end{figure}

\section{Experiment}
\subsection{Dataset and Evaluation Metrics}
We evaluate our method on two publicly available medical datasets: Polyps \cite{Kvasir,CVC-ClinicDB} (\textbf{RGB}) and KiTS19 \cite{KiTS19} (\textbf{CT}, 2D slices), referred to as \textbf{Real Datasets}.

\textbf{Generative Model Training}: For Polyps, we use images from Kvasir \cite{Kvasir} and CVC-ClinicDB \cite{CVC-ClinicDB}. For KiTS19 \cite{KiTS19}, 50 cases are randomly selected from 210 labeled cases, sliced into 2D, filtering out lesion-free slices. 

\textbf{Generative Model Sampling and Evaluation}: Following \cite{ArSDM}, synthetic images are generated using masks from \textbf{Real Datasets}, referred to as \textbf{Synthetic Datasets}, and evaluated using FID \cite{FID} and CLIP-I \cite{CLIP-I}.

\textbf{Segmentation Model Training}: \textbf{Synthetic Datasets} are combined with the \textbf{Real Datasets} as a new training set to train segmentation models.

\textbf{Segmentation Model Testing and Evaluation}: 
The Polyps test set includes images from five public datasets: EndoScene \cite{EndoScene}, CVC-ClinicDB \cite{CVC-ClinicDB}, Kvasir \cite{Kvasir}, CVC-ColonDB \cite{CVC-ColonDB}, and ETIS \cite{ETIS}. For KiTS19 \cite{KiTS19}, 10 non-overlapping cases are selected from 210 labeled cases, sliced into 2D, filtering out slices without lesions. Evaluation metrics include mDice and mIoU for Polyps, and mDice, mIoU, Accuracy, and Recall for KiTS19.

\subsection{Implementation Details}
We detail the configuration of the generative and segmentation models as follows:

\textbf{Generative Model}: We use the pre-trained Stable Diffusion v1.5 \cite{SD}. The training setup is the same for both datasets: the AdamW \cite{AdamW} optimizer with a learning rate of $10^{-5}$ and weight decay of $10^{-2}$ is used for 3,000 iterations on 8$\times$NVIDIA 4090 GPUs (global batch size of 32) with $384^2$ resolution inputs. A 5\% probability for prompt dropout is applied. Sampling employs classifier-free guidance \cite{Classifier-free} (CFG=9) and deterministic DDIM \cite{DDIM} sampling ($\eta=0$, 50 steps), as described in \cite{ControlNet}. T2I-Adapter \cite{T2I} and ControlNet \cite{ControlNet} share the same configuration as our method, while SinGAN \cite{Singan} and ArSDM \cite{ArSDM} use their default settings. Notably, for ControlNet \cite{ControlNet}, unlocking the weights of Stable Diffusion is more effective for medical image synthesis.

\textbf{Segmentation Model}: Both CNN-based and Transformer-based models are utilized with default configurations. Specifically, nnUNet \cite{nnUNet} is trained for 200 epochs with five-fold cross-validation, and the final results are obtained by ensembling five models, followed by postprocessing.

\begin{table*}[t]
\centering
\caption{Comparisons of different methods applied on polyp segmentation baselines.}
\renewcommand\tabcolsep{0.9pt}
{\fontsize{8}{10}\selectfont
\begin{tabular}{@{}lcccccccccc|cc@{}}
\toprule
\multirow{2}{*}{Methods} 
& \multicolumn{2}{c}{EndoScene}                      
& \multicolumn{2}{c}{ClinicDB}     
& \multicolumn{2}{c}{Kvasir}   
& \multicolumn{2}{c}{ColonDB}   
& \multicolumn{2}{c}{ETIS}   
& \multicolumn{2}{|c}{Overall}  
\\ \cmidrule(l){2-13} 
&mDice  &mIoU  &mDice  &mIoU  &mDice  &mIoU  &mDice  &mIoU  &mDice  &mIoU  &mDice  &mIoU        \\             
\midrule
nnUNet & 84.3 & 76.0 & \textbf{89.7} & \textbf{85.0} & 89.7 & 84.3 & 77.2 & 69.2 & 69.1 & 61.5 & 78.3 & 70.9\\
+Copy-Paste & 85.0 & 76.8 & 89.5 & \textbf{85.0} & 89.8 & 84.3 & \textbf{77.7} & \textbf{70.2} & 69.4 & 61.8 & 78.7 & 71.5\\
+SinGAN & 86.5 & 79.4 & 88.8 & 84.0 & 90.2 & 85.4 & 71.7 & 65.7 & 66.7 & 60.5 & 75.2 & 69.3\\
+ArSDM & 86.2 & 79.1 & 89.3 & 84.5 & 90.2 & 84.8 & 75.3 & 68.0 & 73.2 & 65.7 & 78.6 & 71.7\\
+T2I-Adapter & 83.9 & 76.6 & 87.9 & 82.9 & 91.1 & 85.5 & 75.5 & 68.9 & 69.2 & 61.7 & 78.0 & 70.9\\
+ControlNet & 84.2 & 76.5 & 88.6 & 83.8 & 89.9 & 84.5 & 73.6 & 66.0 & 66.6 & 59.1 & 75.9 & 68.8\\
+Ours & \textbf{87.7} & \textbf{79.8} & 88.9 & 84.0 & \textbf{91.3} & \textbf{85.9} & 76.2 & 68.8 & \textbf{74.3} & \textbf{67.8} & \textbf{79.5} & \textbf{72.7}\\
\midrule
SANet & 88.8 & 81.5 & 91.6 & 85.9 & 90.4 & 84.7 & 75.3 & 67.0 & 75.0 & 65.4 & 79.4 & 71.4\\
+Copy-Paste & 89.7 & 83.0 & 90.2 & 85.1 & 90.3 & 84.8 & 77.7 & 70.0 & 77.4 & 68.8 & 81.1 & 73.7\\
+SinGAN & 88.3 & 81.6 & 90.9 & 85.3 & 91.0 & \textbf{85.8} & 77.3 & 69.4 & 73.7 & 65.4 & 80.0 & 72.6\\
+ArSDM & \textbf{90.2} & \textbf{83.2} & 91.4 & 86.1 & 91.1 & 85.6 & 77.7 & 70.0 & 78.0 & 69.5 & 81.5 & 74.1\\
+T2I-Adapter & 89.1 & 81.9 & 91.2 & 85.5 & 90.4 & 84.5 & 77.6 & 70.2 & 76.4 & 67.2 & 81.1 & 73.3\\
+ControlNet & 89.3 & 82.1 & 91.1 & 85.8 & 90.8 & 85.2 & 76.2 & 68.2 & 75.7 & 65.8 & 80.0 & 72.2\\
+Ours & 89.2 & 83.1 & \textbf{92.9} & \textbf{87.4} & \textbf{91.2} & 85.6 & \textbf{77.8} & \textbf{70.4} & \textbf{79.6} & \textbf{71.8} & \textbf{82.0} & \textbf{74.9}\\
\midrule
Polyp-PVT & 90.0 & 83.3 & 93.7 & 88.9 & 91.7 & 86.4 & 80.8 & 72.7 & 78.7 & 70.6 & 83.3 & 76.0 \\
+Copy-Paste &  88.0 & 80.9 & 93.4 & 88.7 & 91.7 & 87.1 & 79.8 & 71.8 & 79.2 & 71.3 & 82.8 & 75.6 \\
+SinGAN & 87.0 & 79.7 & 91.7 & 87.0 & \textbf{92.8} & \textbf{88.1} & 76.9 & 69.0 & 74.2 & 66.7 & 80.1 & 73.0\\
+ArSDM & 88.2 & 81.2 & 92.2 & 87.5 & 91.5 & 86.3 & 81.7 & 73.8 & 80.6 & 72.9 & 84.0 & 76.7\\
+T2I-Adapter & 89.2 & 82.4 & \textbf{94.0} & \textbf{89.2} & 90.4 & 85.0 & 79.6 & 71.7 & 78.1 & 69.8 & 82.4 & 75.1\\
+ControlNet & 86.1 & 78.8 & 91.3 & 85.9 & 91.1 & 86.2 & 79.7 & 71.4 & 78.7 & 70.2 & 82.3 & 74.6 \\
+Ours & \textbf{90.3} & \textbf{83.8} & 93.0 & 88.5 & 92.0 & 87.2 & \textbf{82.0} & \textbf{74.1} & \textbf{80.8} & \textbf{73.1} & \textbf{84.4} & \textbf{77.3}\\
\bottomrule
\end{tabular}}
\label{table3}
\end{table*}

\subsection{Qualitative Comparison}
Fig.~\ref{fig2} demonstrates that the teacher model’s adaptive regularization accelerates the student model’s data fitting within approximately 300 steps, mitigating the sudden convergence phenomenon in ControlNet \cite{ControlNet}.

Fig.~\ref{fig3} and Fig.~\ref{fig4} present kidney tumor and polyp images generated by various methods. SinGAN \cite{Singan}, although designed for the Polyps dataset, often introduces artifacts and lacks diversity. ArSDM \cite{ArSDM} suffers from texture degradation in polyps and fails to generalize to KiTS19 due to its task-specific nature. T2I-Adapter \cite{T2I} generates unrealistic textures in RGB data and underperforms on CT data. ControlNet \cite{ControlNet} struggles with mask-lesion alignment. In contrast, our model excels in both mask-lesion alignment and morphological features, clearly outperforming the others.

\subsection{Quantitative Comparisons}
Table~\ref{table1} shows FID \cite{FID} and CLIP-I \cite{CLIP-I} results. Notably, more precise mask-lesion alignment does not significantly lower the FID score, with our method’s FID score slightly higher than ControlNet \cite{ControlNet}. We attribute this to the inherent limitations of FID \cite{Anomalydiffusion}, which overfits with limited data. Nevertheless, CLIP-I \cite{CLIP-I} confirms our method achieves higher semantic similarity.

Table~\ref{table2} and Table~\ref{table3} highlight the enhancement of segmentation models using synthetic data from various generative models. We establish a new baseline by retraining models on a duplicated dataset (\emph{i.e.}, ``Copy-Paste''). Our method significantly outperforms others. On KiTS19 \cite{KiTS19}, it improves mDice by 2.4\%, mIoU by 4.2\%, and Recall by 2.3\% over TransUNet \cite{Transunet}, and mDice by 1.4\%, mIoU by 2.6\%, and Recall by 1.4\% over nnUNet \cite{nnUNet}. On Polyps, our method outperforms nnUNet \cite{nnUNet} by 1.2\% in mDice and 1.8\% in mIoU, SANet \cite{SANet} by 2.6\% in mDice and 3.5\% in mIoU, and Polyp-PVT \cite{Polyp-PVT} by 1.1\% in mDice and 1.3\% in mIoU. Interestingly, in comparison to ArSDM \cite{ArSDM} and ControlNet \cite{ControlNet}, we observe that there is no consistency between image quality and segmentation performance, indirectly highlighting that our method’s superior mask-lesion alignment is key to improvements across diverse segmentation models.

\begin{table*}[t]
\centering
\caption{Comparison of the impact of $\mathcal{L}_{\text{Ada}}$ on kidney tumor image segmentation.}
\renewcommand\tabcolsep{0.9pt}
{\fontsize{8}{10}\selectfont
\begin{tabular}{@{}lcccc|cccc@{}}
\toprule
\multirow{2}{*}{Settings}
& \multicolumn{4}{c|}{TransUNet} & \multicolumn{4}{c}{nnUNet} 
\\ \cmidrule(l){2-9} 
&mDice &mIoU &Accuracy &Recall &mDice &mIoU &Accuracy &Recall\\
\midrule
w/o & 94.6 & 90.0 & \textbf{99.0} & \textbf{93.9} & 96.1 & 93.2 & \textbf{99.8} & 95.8 \\
w/(Standard) & 94.9 & 90.6 & \textbf{99.0} & 93.5 & 97.4 & 95.3 & 99.6 & 97.6\\
w/(Adaptive) & \textbf{95.2} & \textbf{91.1} & \textbf{99.0} & 93.8 & \textbf{97.9} & \textbf{96.0} & 99.6 & \textbf{97.8} \\
\bottomrule
\end{tabular}}
\label{table4}
\end{table*}

\section{Ablation Study}
We conducted an ablation study to evaluate the importance of the Adaptive Distillation Loss ($\mathcal{L}_{\text{Ada}}$). Table~\ref{table4} presents the results on KiTS19 \cite{KiTS19}. The findings show that regularizing the student model with Distillation Loss (Standard) improves segmentation performance, while $\mathcal{L}_{\text{Ada}}$ (Adaptive) further enhances the baseline model’s accuracy, highlighting its crucial role in mask-lesion alignment.

\section{Conclusion}
We present Adaptively Distilled ControlNet, a novel image synthesis method. During training, a teacher model with image-conditioned inputs adaptively regularizes the student model. During sampling, only the enhanced student model is used, maintaining ControlNet’s \cite{ControlNet} sampling speed. We generate high-quality medical images with accurate mask-lesion alignment and rich morphological features using arbitrary masks. Extensive experiments across two modalities demonstrate the robustness, effectiveness, and superiority of our approach.

\begin{credits}
\subsubsection{\ackname} This work was supported in part by the Startup Grant for Professor (SGP) — CityU SGP, City University of Hong Kong under Grant 9380170.

\end{credits}
\bibliographystyle{splncs04}
\bibliography{main}
\end{document}